\documentclass[10pt,twocolumn,letterpaper]{article}

\usepackage[pagenumbers]{iccv} 
\usepackage[accsupp]{axessibility}  
\usepackage{graphicx}
\usepackage{marvosym}
\usepackage{amsmath}
\usepackage{amssymb}
\usepackage{booktabs}
\usepackage[normalem]{ulem}
\usepackage{bm}
\usepackage{pifont}
\usepackage{multirow}
\usepackage{colortbl}
\usepackage{pgfplots}
\definecolor{mygray}{gray}{.9}
\useunder{\uline}{\ul}{}

\newcommand{\re}[2]{\textcolor{#1}{#2}}
\definecolor{iccvblue}{rgb}{0.21,0.49,0.74}
\usepackage[pagebackref,breaklinks,colorlinks,allcolors=iccvblue]{hyperref}

\def\paperID{2428} 
\def\confName{ICCV}
\def\confYear{2025}

\title{CleanPose: Category-Level Object Pose Estimation via Causal Learning and Knowledge Distillation}

\author{
    Xiao Lin\textsuperscript{1}\hspace{1.5em} 
    Yun Peng\textsuperscript{1}\hspace{1.5em} 
    Liuyi Wang\textsuperscript{1}\hspace{1.5em} 
    Xianyou Zhong\textsuperscript{1}\hspace{1.5em}  
    Minghao Zhu\textsuperscript{1} \\
    Yi Feng\textsuperscript{1}\hspace{1.5em} 
    Jingwei Yang\textsuperscript{1}\hspace{1.5em} 
    Chengju Liu\textsuperscript{1,2}\thanks{Corresponding authors.}\hspace{1.5em} 
    Qijun Chen\textsuperscript{1,2}\footnotemark[1]\\ 
\textsuperscript{1}College of Electronic and Information Engineering, Tongji University, Shanghai, China\\
\textsuperscript{2}State Key Laboratory of Autonomous Intelligent Unmanned Systems\\
{\tt\small \{linx\_xx, pengyun, wly, zhongxianyou, zmhh\_h, jw\_yang,}\\
{\tt\small  fengyi0109, liuchengju, qjchen\}@tongji.edu.cn}
}

\begin{document}
\maketitle
\begin{abstract}    
    In the effort to achieve robust and generalizable category-level object pose estimation, recent methods primarily focus on learning fundamental representations from data. However, the inherent biases within the data are often overlooked: the repeated training samples and similar environments may mislead the models to over-rely on specific patterns, hindering models' performance on novel instances.    
    In this paper, we present \textbf{CleanPose}, a novel method that mitigates the data biases to enhance category-level pose estimation by integrating causal learning and knowledge distillation.
    By incorporating key causal variables (structural information and hidden confounders) into causal modeling, we propose the causal inference module based on front-door adjustment, which promotes unbiased estimation by reducing potential spurious correlations.
    Additionally, to further confront the data bias at the feature level, we devise a residual-based knowledge distillation approach to transfer unbiased semantic knowledge from 3D foundation model, providing comprehensive causal supervision.
    Extensive experiments across multiple benchmarks (REAL275, CAMERA25 and HouseCat6D) hightlight the superiority of proposed CleanPose over state-of-the-art methods. 
    Code will be available at \url{https://github.com/chrislin0621/CleanPose}.
\end{abstract}    
\section{Introduction}
\label{sec:intro}

\begin{figure}[htbp]
\centering
\includegraphics[width=.95\columnwidth]{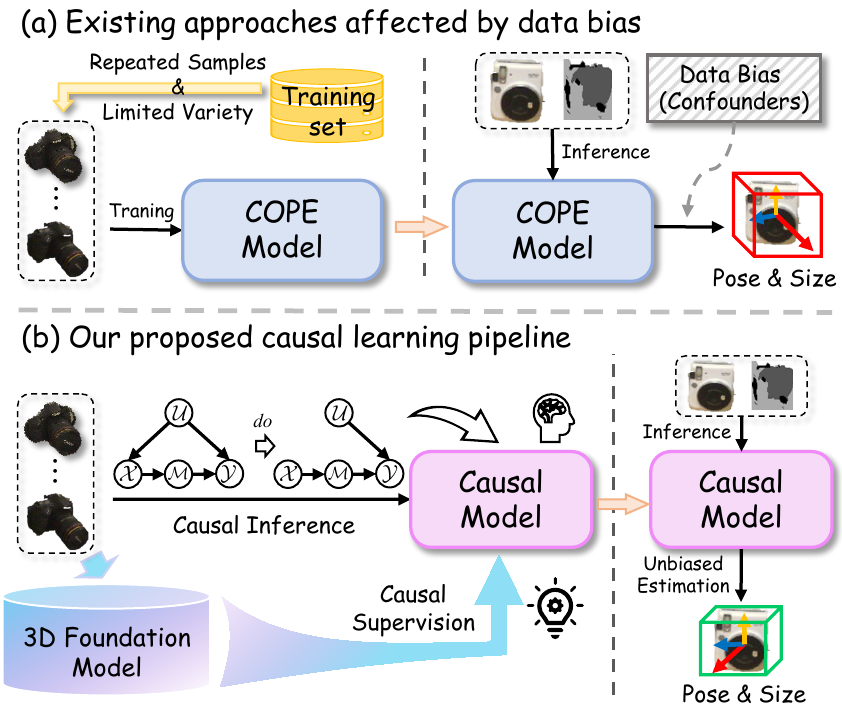}
   \caption{Comparison of (a) existing pose estimation approaches and (b) the proposed causal learning pipeline. The causal inference is employed to mitigate the negative effect raised by bias, and 3D semantic knowledge distilled from foundation model is leveraged to provide comprehensive causal guidance.
   }
   \vspace{-0.4cm}
   \label{fig:head_fig}
\end{figure}


\textbf{C}ategory-level \textbf{o}bject \textbf{p}ose \textbf{e}stimation (COPE)~\cite{wang2019normalized} aims to predict the 9Dof pose for arbitrary objects within predefined categories. 
This task, unlike instance-level pose estimation~\cite{wang2019densefusion,zhou2021semi,he2021ffb6d,hai2023shape,lin2024transpose}, does not require high-quality CAD models, making it feasible to perceive a broader range of novel objects instead of a single instance. 
The complex intra-category variations have long been regarded as main challenge of COPE, thus recent methods~\cite{chen2021sgpa,li2025gce,di2022gpv,lin2022category,zheng2023hs,lin2023vi,lin2024clipose,lin2024instance,chen2024secondpose} mainly focus on model designing to capture robust category features from data. 
Despite remarkable development, the performance of COPE models remains limited~\cite{lin2024clipose,lin2023vi}, suggesting that feature learning may not be the fundamental issue.
This prompts us to identify the key hindrance blocking further progress.
We observe that inherent biases are pervasive in current datasets~\cite{wang2019normalized}, \eg, repeated training samples and limited pose variety, which may mislead COPE models to overfit to familiar object's appearance and poses during data fitting, as shown in \cref{fig:head_fig}(a).
However, existing efforts often overlook these underlying data biases and the resulting negative impacts.

One way to tackle bias issue is to create broader and more diverse datasets~\cite{fu2022category,wang2022phocal,jung2024housecat6d,zhang2024omni6d}. Though valuable, achieving a perfectly balanced dataset free of bias remains nearly impossible. Additionally, the dataset's scale is still significantly constrained by the cost of 3D data annotation~\cite{zhang2024pcp}. Therefore, the extension of datasets does not fundamentally solve the hindrance, developing a causal COPE models that can effectively confront and alleviate biases becomes a primary challenge.

To address this issue, we direct our attention to the human observation mechanism. In fact, the reason why humans can effectively handle variations among intra-category objects is that we can leverage the analogical association to infer the structural features and underlying causal relations.
Moreover, the observations of numerous objects enable humans to maintain a stable perception of novel objects across different environments and viewpoints.

Motivated by this finding, for the first time, we propose to incorporate the causal inference~\cite{pearl2009causality} into the formulation of COPE, exploiting the concepts of \emph{intervention}~\cite{pearl2016causal} to mitigate the negative effects caused by confounders. Here, confounders are variables that influence both input and outcomes, just as dataset biases in this task simultaneously affect input sampling and output pose. The causal learning can investigate the causal relation among variables and equip COPE models with similar cognitive abilities that humans have.
However, it is non-trivial to directly incorporate the causal inference in COPE applications because of the following challenges: (\textbf{i}) The unique modality of 3D data makes it impractical to directly apply existing causal modeling techniques~\cite{huang2024causalpc}. (\textbf{ii}) Subsequently, the confounders in pose estimation task are inherently unobserved and elusive, which further increases the challenge of identifying these confounders. (\textbf{iii}) Moreover, limited training samples raise the difficulties in learning comprehensive causal relations.

To overcome above challenges, we present \textbf{CleanPose}, a concise yet effective framework with \textbf{c}ausal \textbf{l}earning and knowledg\textbf{e} distill\textbf{a}tio\textbf{n} to enhance the category-level \textbf{pose} estimation, as shown in \cref{fig:head_fig}(b).
Faced with the challenge that hidden confounders are elusive or even unobserved in our task, we have two key designs: (\textbf{i}) The first design is a causal inference module based on \emph{front-door adjustment}, which can effectively approximate the predicted expectations by potential confounders.
Moreover, to suitably represent confounders, we devise and maintain a dynamic queue to efficiently update training samples, similar with MoCo~\cite{he2020momentum}.
To further mitigate biases, (\textbf{ii}) our second design is a residual-based feature knowledge distillation module, which can transfer unbiased point cloud category information of the 3D foundation model, ULIP-2~\cite{xue2024ulip}, into our model. Since the foundation model is pretrained on a diverse range of objects, it captures robust semantic features that can implicitly provide comprehensive causal supervision at the feature level.
As demonstrated by extensive experiments, our findings reveal the impact of integrating causal learning to reduce biases, and providing comprehensive causal supervision, enhancing the model's robustness and generalization.

To summarize, our main contributions are as follows:
\begin{itemize}
    \item We propose CleanPose, the first solution to mitigate the confoundering effect in category-level pose estimation via causal learning. Taking inspiration from human observation mechanism, we propose to identify the causal effect to achieve unbiased estimation, recovering correct pose of novel instances within predefined categories.
    \item To provide comprehensive causal supervision and enhance intra-category generalization, we develop a residual category knowledge distillation approach to transfer rich and unbiased 3D semantic knowledge from 3D foundation models into COPE networks.
    \item Our proposed CleanPose achieves state-of-the-art performance on three mainstream challenge benchmarks (REAL275~\cite{wang2019normalized}, CAMERA25~\cite{wang2019normalized} and HouseCat6D~\cite{jung2024housecat6d}). For instance, the accuracy attains \textbf{61.7\%} on rigorous metric 5°2 \emph{cm} of REAL275 dataset, surpassing the current best method with a large margin by 4.7\%.
\end{itemize}

\section{Related Works}
\label{sec:relate_works}
\noindent
{\bf Category-level Object Pose Estimation.} The objective of this task encompasses predicting the 9DoF pose for unseen objects within predefined categories.
To address this challenging task, pioneer method NOCS~\cite{wang2019normalized} suggests mapping input shape to a normalized canonical space and recovering the pose via Umeyama algorithm~\cite{umeyama1991least}. 
SPD~\cite{tian2020shape} proposes a method for deriving and utilizing the shape prior for each category. This crucial insight inspires many subsequent prior-based works~\cite{lin2022sar,fan2021acr,li2025gce}, which further improve the use of shape priors, continuously improving the pose estimation performance.
More recently, prior-free methods~\cite{di2022gpv,zheng2023hs,lin2023vi,chen2024secondpose,lin2024instance} have achieved impressive performance.
VI-Net~\cite{lin2023vi} separates rotation into viewpoint and in-plane rotations, while SecondPose~\cite{chen2024secondpose} propose to extract hierarchical panel-based geometric features for point cloud.
AG-Pose~\cite{lin2024instance} achieves current state-of-the-art performance by explicitly extract local and global geometric keypoint information of different instances.
However, above methods have not fully considered the dataset biases behind the task, which may leads the models to learn spurious correlations, damaging the generalization ability to unseen instances within the categories.

\begin{figure}[htbp]
\centering
\includegraphics[width=.95\columnwidth]{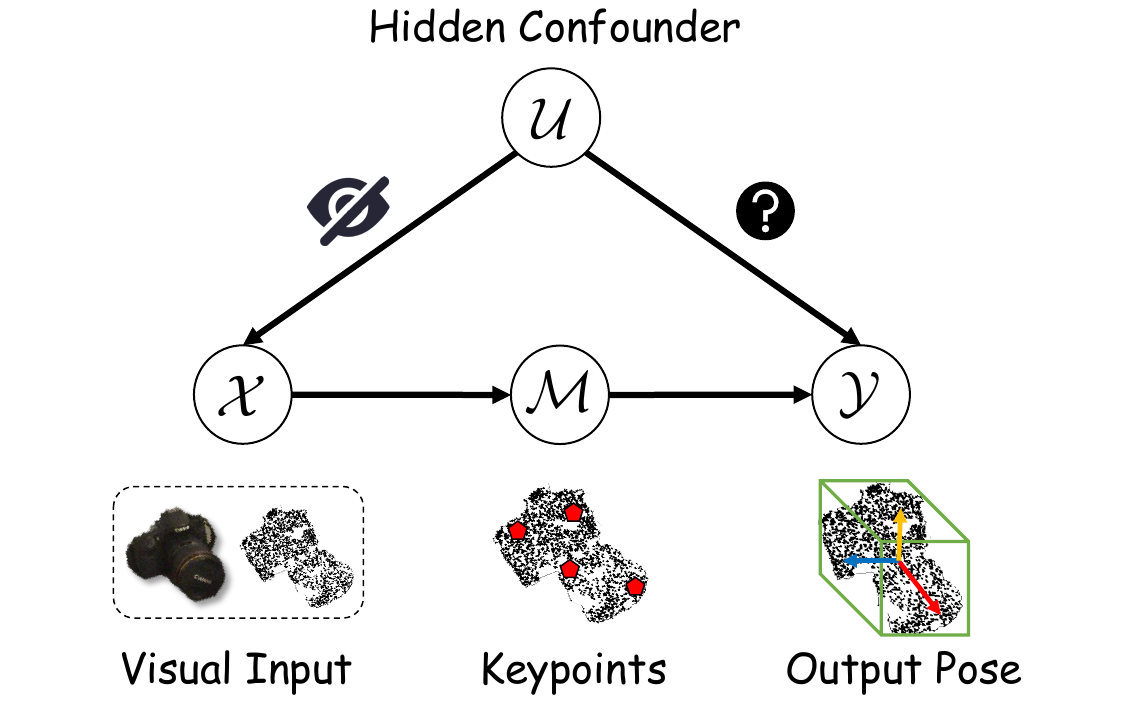}
   \caption{Illustration of the structural causal model of COPE.
   }
   \vspace{-0.4cm}
   \label{fig:causal_model}
\end{figure}

\vspace{0.05cm}
\noindent
{\bf Causal Inference.}
Causality~\cite{pearl2009causality} is an emerging technique refering to the modeling the relationships between factors in a task from a human perspective.
Recent methods have incorporated causal inference to improve the performance of DNNs in computer vision domains, \eg, object detection~\cite{zhang2022multiple,wang2021causal}, image captioning~\cite{yang2021deconfounded,liu2022show} and vision-language task~\cite{yang2021causal,zhang2024if,wang2024vision}.
Though valuable, these methods can flexibly identify confounders due to natural prompts in data, \eg, the keywords in instructions~\cite{wang2024vision} and salient regions in an image~\cite{wang2021causal}, which are inapplicable for point cloud data.

In 3D domain, several works have focused on enhancing the robustness of point cloud classification~\cite{huang2024causalpc} or 3D reconstruction~\cite{liu2022structural} with causal inference.
However, it is non-trivial to adapt these approaches to pose estimation due to the inherent differences in human modeling among these tasks.
In this work, we propose a specific causal learning approach based on front-door adjustment for category-level pose estimation for the first time.

\vspace{0.05cm}
\noindent
{\bf Knowledge Distillation.} Knowledge distillation~\cite{hinton2015distilling} is a technique that transfers knowledge from a teacher model to a student model. Transferring the knowledge of foundation models for downstream tasks has been proven to be effective~\cite{gu2024open,huang2024froster,zhu2024mote}. ViLD~\cite{gu2024open} distill knowledge from CLIP~\cite{radford2021learning} to achieve open-vocabulary object detection, while some methods~\cite{huang2024froster,zhu2024mote} utilize the generalizability of VLMs to address the video recognition task. 
In this work, we leverage the 3D foundation model ULIP-2~\cite{xue2024ulip} as a teacher model to guide the student model in correcting biases at the feature level, enhancing the robustness and generalization capabilities of COPE model from multiple perspectives.



\section{Preliminaries}
\label{sec:Preliminary}
\subsection{Task Formulation}
\label{sec:task_formulation}
Given an RGB-D image containing objects from a predefined set of categories, off-the-shelf segmentation models such as MaskRCNN~\cite{he2017mask} are employed to obtain masks and category labels for each object in a image. Then, the segmentation masks can be utilized to get the cropped RGB image $\mathcal{I}_{obj} \in \mathbb{R}^{H \times W \times 3}$ and the point cloud $\mathcal{P}_{obj} \in \mathbb{R}^{N \times 3}$, where $N$ is the number of points and $\mathcal{P}_{obj}$ is acquired by back-projecting the cropped depth image with camera intrinsics followed by a downsampling process. With the input $\mathcal{I}_{obj}$ and $\mathcal{P}_{obj}$, the objective of COPE~\cite{wang2019normalized} is to recover the 9DoF poses, including the 3D rotation $\mathcal{R} \in SO(3)$, the 3D translation $t \in \mathbb{R}^3$, and 3D metric size $s \in \mathbb{R}^3$.


\subsection{The Causal Modeling of CleanPose}
\label{sec:casual_modeling}
To quantify the underlying logic behind human observation, we construct a structural causal model~\cite{pearl2009causality,pearl2016causal} capturing the relationship among the key variables in COPE: visual input $\mathcal{X}$, output pose $\mathcal{Y}$, mediator $\mathcal{M}$ and hidden confounders $\mathcal{U}$.
We illustrate the causal model in \cref{fig:causal_model}, where each direct link denotes a causal relationship between two nodes.
\begin{itemize}
    \item $\mathcal{X}\rightarrow\mathcal{M}\rightarrow\mathcal{Y}$ (\emph{Front-door path}): Typically, humans first recognize the structural information of an object, \ie, keypoints~\cite{lin2024instance,zheng2023hs}, and then determine the object's pose based on the similar poses of other objects within this category. This process involves identifying keypoints of the object and their relative positions, and leveraging this information to perform pose estimation. We use mediator $\mathcal{M}$ to represent such structural information and describe such process via the causal path $\mathcal{X} \rightarrow \mathcal{M} \rightarrow \mathcal{Y}$, which is  also referred to as the \emph{front-door path}. 
    \item $\mathcal{X} \leftarrow \mathcal{U} \rightarrow \mathcal{Y}$ (\emph{Hidden confounders}): The confounders are extraneous variables that influence both inputs and outputs, \eg, dataset biases~\cite{jung2024housecat6d,zhang2024omni6d} or category specific attributes~\cite{tian2020shape}. $\mathcal{U} \rightarrow \mathcal{X}$ exists because the input data is inevitably affected by the limited resources in real world and sampling noises from sensors when collection and simulation. Moreover, $\mathcal{U} \rightarrow \mathcal{Y}$ emerges because collected scenes, annotation bias, or the variety of pose also affect the probability of pose distributions. Existing COPE methods tend to model the statistical correlations $P(\mathcal{Y}|\mathcal{X})$, given the optimization goal of maximizing the pose accuracy~\cite{zheng2024georef}. With the lack of modeling of hidden confounders, no matter how large the amount of training data is, the model can not identify the true causal effect from $\mathcal{X}$ to $\mathcal{Y}$.
\end{itemize}

\section{Methodology}
\label{sec:Methodology}

\begin{figure*}[htbp]
    \includegraphics[width=\textwidth]{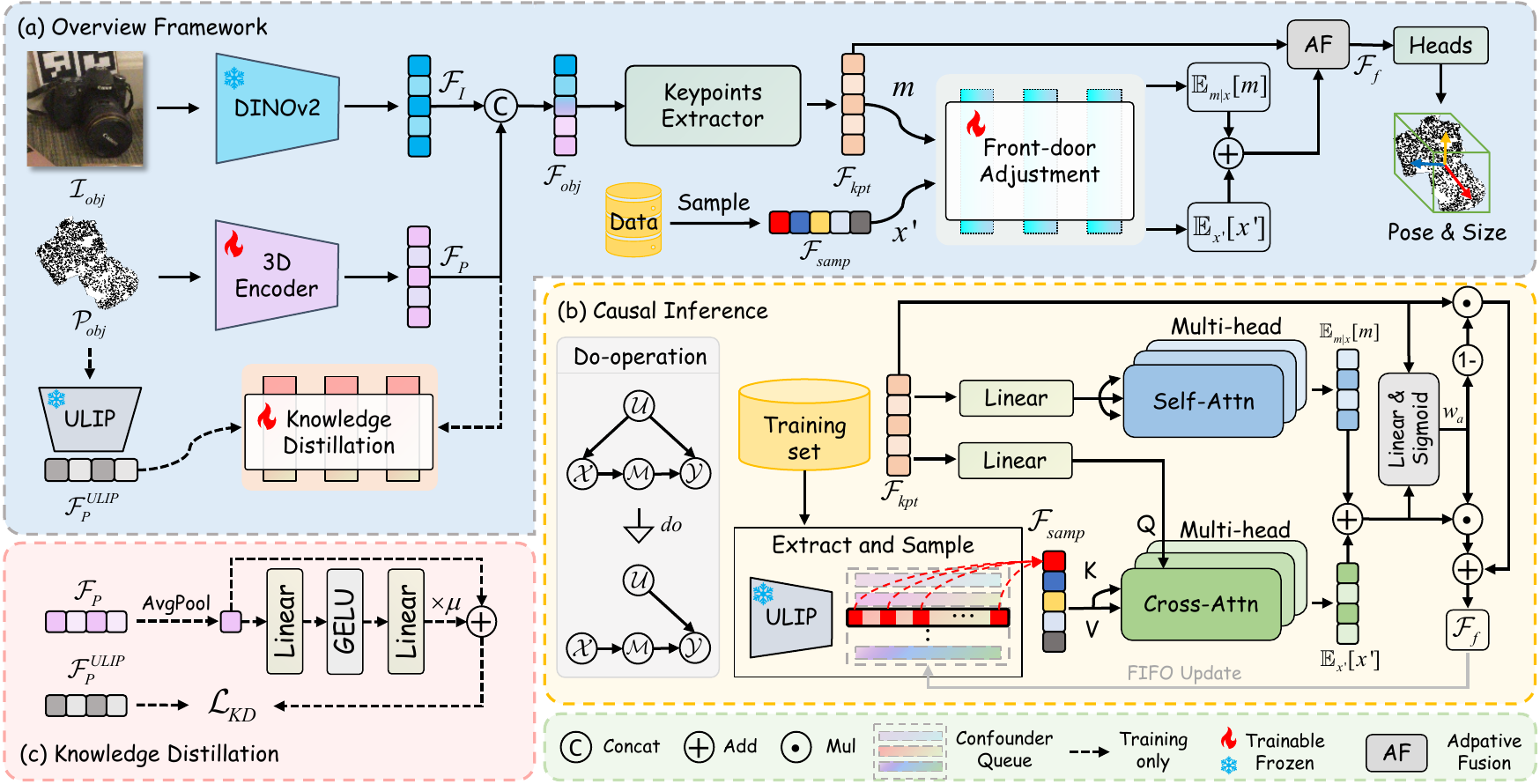}
    \caption{(a) Framework of CleanPose. (b) Causal inference based on front-door adjustment is employed for mitigating potential spurious correlations, promoting unbiased pose estimation. Moreover, the proposed (c) residual-based knowledge distillation module can efficiently provide comprehensive category-specific guidance.
    }\label{fig:CleanPose}
    \vspace{-0.4cm} 
\end{figure*}

\subsection{Feature Extractor}
\label{sec:feature_extractor}
Following~\cite{lin2024instance}, we utilize the PointNet++~\cite{qi2017pointnet++} to extract point feature $\mathcal{F}_{P} \in \mathbb{R}^{N \times C_1}$ of input point cloud $\mathcal{P}_{obj}$.
As for the RGB image $\mathcal{I}_{obj}$, we adopt DINOv2 (ViT-S/14)~\cite{oquab2024dinov2} as our image feature extractor, which has been proven to extract abundant semantic-aware information from RGB images~\cite{peng2024sam}. We select those pixel features corresponding to $\mathcal{P}_{obj}$ and utilize linear interpolation to propagate the original DINOv2 features into the final RGB features $\mathcal{F}_{I} \in \mathbb{R}^{N \times C_2}$. Eventually, we concatenate $\mathcal{F}_{P}$ and $\mathcal{F}_{I}$ to form $\mathcal{F}_{obj} \in \mathbb{R}^{N \times C}$ as the input for the subsequent networks.

\subsection{Causal Inference of CleanPose}
\label{sec:causal_inference}
{\bf The Adjustment Formulation.} 
To avoid making the pose estimation process overfit to specific hidden confounders, we propose to perform causal modeling according to the causal graph and the concepts of \emph{intervention}~\cite{pearl2009causality}.
Specifically, the process of \emph{intervention} can be reflected as \emph{do}-operation~\cite{pearl2016causal}, which provides scientifically sound methods for determining causal effects, as shown in \cref{fig:CleanPose}(b). The specific casual modeling via \emph{do}-operation is as follow.
Based on Bayes's theorem, the typical observational likelihood is as:
\begin{small}
\begin{equation}
\label{equ:typical_likelihood}
    P(\mathcal{Y}|\mathcal{X}) = \sum_{u}P(\mathcal{Y}|\mathcal{X},u)P(u|\mathcal{X}),
\end{equation}
\end{small}
where $P(u|\mathcal{X})$ would bring biased weights. Then we perform the \emph{do}-operation on inputs $\mathcal{X}$, which amounts to removing all edges directed into that variable in the graph model, \ie, from hidden confounders $\mathcal{U}$. According to the invariance and independence rules~\cite{pearl2016causal}, we have:
\begin{small}
\begin{align}
    P(\mathcal{Y}|do(\mathcal{X})) &= \sum_{u}P(\mathcal{Y}|do(\mathcal{X}),u)P(u|do(\mathcal{X})) \label{equ:typical_likelihood_dox_1}\\
    &= \sum_{u}P(\mathcal{Y}|\mathcal{X},u)P(u) \label{equ:typical_likelihood_dox_2}.
\end{align}
\end{small}
In this way, the intervention is realized by blocking the causal path $\mathcal{U} \rightarrow \mathcal{X}$. The \cref{equ:typical_likelihood_dox_2} is also known as the \emph{adjustment formulation}~\cite{pearl2009causality}.

\vspace{0.1cm}
\noindent
{\bf Front-door Adjustment Causal Learning.} In the previous section, we have clarified meaning of the \emph{front-door path} $\mathcal{X} \rightarrow \mathcal{M} \rightarrow \mathcal{Y}$. In COPE task, the DNNs-based models $P(\mathcal{Y}|\mathcal{X}) = \sum_{m}P(\mathcal{Y}|m)P(m|\mathcal{X})$ will choose the suitable knowledge $\mathcal{M}$ from input $\mathcal{X}$ for pose estimation results $\mathcal{Y}$.
Due to the presence of the mediators $\mathcal{M}$, applying adjustment \cref{equ:typical_likelihood_dox_2} solely to the inputs $\mathcal{X}$ is insufficient to fully block the confounding effects~\cite{wang2024vision}. Therefore, we implement continuous intervention in both stages of front-door path~\cite{pearl2016causal}, \ie, by simultaneously applying \emph{do}-operation to $\mathcal{X}$ and $\mathcal{M}$.
First, the effect of $\mathcal{X}$ on $\mathcal{M}$ is identifiable:
\begin{small}
\begin{equation}
\label{equ:x2m_do}
    P(\mathcal{M}|do(\mathcal{X})) = P(\mathcal{M}|\mathcal{X}).
\end{equation}
\end{small}
Then, for the second stage $\mathcal{M} \rightarrow \mathcal{Y}$, we apply the adjustment formulation and have:
\begin{small}
\begin{equation}
\label{equ:m2y_do}
    P(\mathcal{Y}|do(\mathcal{M})) = \sum_{x'}P(\mathcal{Y}|\mathcal{M}, x')P(x').
\end{equation}
\end{small}
Here, $x'$ denotes potential inputs of the whole representation space, different from current inputs $\mathcal{X} = x$. To chain together these two partial effects to obtain the overall effect of $\mathcal{X}$ on $\mathcal{Y}$, we sum over all states $m$ of $\mathcal{M}$ to form overall front-door adjustment as follow:
\begin{small}
\begin{align}
    P(\mathcal{Y}|do(\mathcal{X})) &= \sum_{m}P(\mathcal{Y}|do(m))P(m|do(\mathcal{X})) \label{equ:x2y_do_1} \\
    &= \sum_{x'}P(x')\sum_{m}P(\mathcal{Y}|m, x')P(m|\mathcal{X}) \label{equ:x2y_do_2}.
\end{align}
\end{small}
From the above derivation, we can observe that the front-door adjustment in causal inference involves the intermediate knowledge of current input $\mathcal{X} = x$ (\ie, the current features) as well as the cross-sampling features of other samples from the entire training set. We utilize the bold symbol $\boldsymbol{m}$ and $\boldsymbol{x'}$ to represent these two components respectively. Here, the intermediate knowledge is extracted from current input $\boldsymbol{m} \sim x$. Detailed methods for sampling $\boldsymbol{x'}$ are introduced in subsequent sections.

For convenience, we define the network module as a linear function $f(x,\boldsymbol{x'})$ to model the computation of $x$ and $\boldsymbol{x'}$ to obtain $\mathcal{Y}$. Meanwhile, according to the Bayes' rule and the definition of Expected values, the front-door adjustment $\mathcal{FD}$(·) of \cref{equ:x2y_do_2} can be expressed as:
\begin{small} 
\begin{align}
    P(\mathcal{Y}|do(\mathcal{X})) &= \mathcal{FD}(x,\boldsymbol{x}') \label{equ:x2y_do_expected_1}\\
    &= \mathbb{E}_{x'}\mathbb{E}_{m|x}[f(x,\boldsymbol{x'})] \label{equ:x2y_do_expected_2},
\end{align}
\end{small}
where the linear function is used as $f(x,\boldsymbol{x'}) = f_x(x) + f_{x'}(x')$. Hence, based on the linear mapping model, the \cref{equ:x2y_do_expected_2} can be formulated as $\mathbb{E}_{x'}[\boldsymbol{x'}] + \mathbb{E}_{m|x}[\boldsymbol{m}]$. Since the closed-form solution of expected values are difficult to obtain in the complex representation space~\cite{wang2024vision}, we employ the query mechanism to achieve the estimation:
\begin{small} 
\begin{align}
    \mathbb{E}_{x'}[\boldsymbol{x}'] &\approx \sum_{x'}P(\boldsymbol{x}'|\boldsymbol{g}_1)\boldsymbol{x}' = \sum_{i} \frac{\exp \left(\boldsymbol{g}_{1} \boldsymbol{x}_{i}^{\prime T}\right)}{\sum_{j} \exp \left(\boldsymbol{g}_{1} \boldsymbol{x}_{j}^{\prime T}\right)} \boldsymbol{x}_{i}^{\prime} \label{equ:query1} \\
    \mathbb{E}_{m|x}[\boldsymbol{m}] &\approx \sum_{m}P(\boldsymbol{m}|\boldsymbol{g}_2)\boldsymbol{m} = \sum_{i} \frac{\exp \left(\boldsymbol{g}_{2} \boldsymbol{m}_{i}^{T}\right)}{\sum_{j} \exp \left(\boldsymbol{g}_{2} \boldsymbol{m}_{j}^{T}\right)} \boldsymbol{m}_{i} \label{equ:query2},
\end{align}
\end{small}
where $\boldsymbol{g}_1 = q_1(x)$ and $\boldsymbol{g}_2 = q_2(x)$ are two embedding functions~\cite{yang2021deconfounded,liu2023cross} that transmit input $x$ into two query sets. Then, the front-door adjustment $\mathcal{FD}$(·) is approximated as follows:
\begin{small} 
\begin{equation}
\label{equ:fd_final}
    \mathcal{FD}(x,\boldsymbol{x}') = \mathbb{E}_{x'}[\boldsymbol{x}'] + \mathbb{E}_{m|x}[\boldsymbol{m}].
\end{equation}
\end{small}
As we realize an approximate estimation through a querying mechanism, multi-head attention~\cite{vaswani2017attention} can be efficiently employed to handle the aforementioned process.

\vspace{0.2cm}
\noindent
{\bf Specific Network Design.} Although the derivation is clear, how to implement the causal inference in networks presents a significant technical challenge. 
Considering the characteristics of COPE, we propose leveraging the keypoint features of object as the intermediate knowledge $\boldsymbol{m}$, since the keypoints that are evenly distributed on the surface can effectively capture structure and pose information of the object.
Specifically, with the fusion input feature $\mathcal{F}_{obj}$, we detect $N_{kpt}$ local keypoints and perform global information aggregation to extract features $\mathcal{F}_{kpt} \in \mathbb{R}^{N_{kpt} \times C}$, following previous keypoint-based methods~\cite{lin2024instance,liu2023net}.
As for the cross-sampling features $\boldsymbol{x'}$, one straightforward way is to construct a memory bank of all features and randomly sample a certain number of features from it as the cross-sampling features. Afterward, the memory bank is updated at the corresponding positions using the features from the current mini-batch~\cite{wu2018unsupervised}. However, this intuitive approach is not applicable to our task, since the memory bank should cover all samples, it may not capture the dynamic variations of features during training. The lack of feature consistency could introduce additional confounders, which will negatively impact the models to focus on correct causal relations.

To this end, we draw inspiration from MoCo~\cite{he2020momentum} and devise a features sampling approach based on a dynamic queue. 
Specifically, we first construct a 2D queue with a shape of $N_{c} \times N_{q}$, where $N_{c}$ and $N_{q}$ represent the number of categories and the queue length of each category, respectively. 
We then select training instances at random equal to $N_{q}$ for each category, extract their features using 3D encoders of ULIP-2~\cite{xue2024ulip}, and store them into the queue to complete the initialization.
Subsequently, we randomly sample $N_{s}$ features from the queue to form $\boldsymbol{x'}$, which is denoted as $\mathcal{F}_{samp} \in \mathbb{R}^{N_{s} \times C}$ in the networks. 
To maintain feature consistency within the queue, we update it using the output features $\mathcal{F}_{f}$ from the current epoch using a first-in-first-out (FIFO) strategy.
Following \cref{equ:query1} to \cref{equ:fd_final}, the causality-enhanced features $\mathcal{F}_{f}$ are obtained as follow:
\begin{small} 
\begin{equation}
\label{equ:fd_attn}
    \mathcal{F}_{s} = SA(\mathcal{F}_{kpt}), \mathcal{F}_{c} = CA(\mathcal{F}_{kpt}, \mathcal{F}_{samp})
\end{equation}
\begin{equation}
\label{equ:fd_ln_plus}
    \mathcal{F}_{f} = LN(\mathcal{F}_{s} + \mathcal{F}_{c}),
\end{equation}
\end{small}
where $SA$(·) and $CA$(·) represent multi-head self-attention and cross-attention, respectively. $LN$(·) denotes the layer normalization.
Furthermore, to strengthen the stability of learning by fusing causality-enhanced features with original keypoint features of objects, we introduce an adaptive weight fusion method:
\begin{small} 
\begin{equation}
\label{equ:gate1}
    w_{a} = \sigma(\mathcal{F}_{f}W_{f}+\mathcal{F}_{kpt}W_{k})
\end{equation}
\begin{equation}
\label{equ:gate2}
    \mathcal{F}_{f} \leftarrow w_{a}\odot \mathcal{F}_{f} + (1-w_{a})\odot \mathcal{F}_{kpt},
\end{equation}
\end{small}
where $\sigma$ and $\odot$ mean Sigmoid function and element-wise multiplication. Suppose $W_{f/k} \in \mathbb{R}^{C \times 1}$ is learnable weight parameter.

\subsection{Knowledge Distillation}
\label{sec:knowledge_distillation}
As mentioned in above section, the foundation model is trained on a broader dataset covering a wide range of categories and scenarios, enabling it to be exposed to diverse data distributions and to implicitly learn more robust debiased feature representations, which can supervise the COPE model to confront biases at the feature level.
Inspired by the feature alignment design~\cite{huang2024froster,zhu2024mote,zhu2023fine} to transfer visual-language knowledge to different tasks, we exploit 3D foundation model ULIP-2~\cite{xue2024ulip} as the teacher model to provide causal guidance for COPE model.
Specifically, we introduce a residual-based approach to distill debiasing knowledge from 3D foundation models into COPE network, as shown in \cref{fig:CleanPose}(c).
Unlike CLIPose~\cite{lin2024clipose} exploits contrastive learning to extract semantic knowledge from text and image modalities, we leverage the pre-trained 3D encoder of ULIP-2, which possesses strong awareness of point cloud structures and category information.
Specifically, given the input point cloud of a object $\mathcal{P}_{obj}$, we can directly feed it into the frozen 3D Encoder $\Phi_{ULIP}$ of ULIP-2, which can be written as:
\begin{small} 
\begin{equation}
\label{equ:pointbert}
    \mathcal{F}^{ULIP}_{P} = \Phi_{ULIP}(\mathcal{P}_{obj}),
\end{equation}
\end{small}
where $\mathcal{F}^{ULIP}_{P} \in \mathbb{R}^{1 \times C_3}$ denotes the $\verb|[CLS]|$ token in Point Transformer. The feature of $\verb|[CLS]|$ token embeds representation of the whole point cloud. To obtain the representation of our model, we employ a simple average pooling on point features and get $\mathcal{F}_{P}^{avg} \in \mathbb{R}^{1 \times C_1}$, we have:
\begin{small} 
\begin{equation}
\label{equ:avg_pool}
    \mathcal{F}_{P}^{avg} = \mathrm{AvgPool}(\mathcal{F}_{P}).
\end{equation}
\end{small}
Then, we apply a modified residual distillation network on $\mathcal{F}_{P}^{avg}$ to transform with two MLP projectors $K_1$ and $K_2$:
\begin{small} 
\begin{equation}
\label{equ:residual}
    \mathcal{\widehat{F}}_{P}^{avg} = \mathcal{F}_{P}^{avg} + \mu \times K_{2}(\delta(K_{1}(\mathcal{F}_{P}^{avg}))),
\end{equation}
\end{small}
where $K_{1/2} \in \mathbb{R}^{C_1 \times C_1}$, $\delta$ represents GeLU function and $\mu$ is a balancing parameter. Moreover, we initialize the parameters of the second layer $K_{2}$ as zeros. Therefore, $\mathcal{\widehat{F}}_{P}^{avg}$ initially contains only $\mathcal{F}_{P}^{avg}$ and is gradually updated, which avoids introducing additional confounders.
Lastly, we can utilize L2 loss to supervise the category knowledge distillation:
\begin{small} 
\begin{equation}
\label{equ:distillation}
    \mathcal{L}_{KD} = \frac{1}{B} \sum_{i}^{B}\left\|\mathcal{F}^{ULIP}_{P}-\psi(\mathcal{\widehat{F}}_{P}^{avg})\right\|_{2},
\end{equation}
\end{small}
where $\psi \in \mathbb{R}^{C_1 \times C_3}$ is a learnable embedding layer and $B$ denotes the batch size. 
In this way, we effectively transfer the debiasing knowledge from the foundation model to the COPE model, further mitigating the interference of data biases.
Moreover, the tuned point cloud features of our model can effectively receive supervision from generalized ones, facilitating comprehensive category knowledge learning and enhancing the model's intra-category generalization.

\subsection{Pose Estimation and Overall Loss Function}
\label{sec:estimation_and_loss}
With the obtained causality-enhanced features, following previous works~\cite{lin2022category,lin2024instance}, we recover the final pose and size $\mathcal{R}, t, s$ via set of keypoint-level correspondences containing global features and points. A simply L1 Loss is used to supervise the predicted pose, in formula:
\begin{small} 
\begin{equation}
\label{equ:pose_rts}
    \mathcal{L}_{pose} = \left\|\mathcal{R}_{gt} - \mathcal{R}\right\|_{2} + \left\|t_{gt} - t\right\|_{2} + \left\|s_{gt} - s\right\|_{2},
\end{equation}
\end{small}
where $\mathcal{R}_{gt}, t_{gt}, s_{gt}$ means the ground truth rotation, translation and size. For more details please refer to supplementary or ~\cite{lin2024instance}. 
Hence, the overall loss function is as follow:
\begin{small} 
\begin{equation}
\label{equ:overall_loss_func}
    \mathcal{L}_{all} = \alpha_{1}\mathcal{L}_{pose} + \alpha_{2}\mathcal{L}_{KD},
\end{equation}
\end{small}
where $\alpha_{1}, \alpha_{2}$ are hyper-parameters to balanced the contribution of each term. We omit some loss terms for brevity. Please refer to appendix for more details.

\section{Experiments}
\label{sec:Experiments}

{\bf Datasets.} Following previous works~\cite{lin2023vi,lin2024instance,lin2024clipose,zheng2024georef,chen2024secondpose}, we conduct experiments not only on two mainstream NOCS benchmarks, REAL275~\cite{wang2019normalized} and CAMERA25~\cite{wang2019normalized} as well as HouseCat6D~\cite{jung2024housecat6d} datasets.
REAL275 is a challenge real-world dataset that contains objects from 6 categories. The training data consists of 4.3k images from 7 scenes, while testing data includes 2.75k from 6 scenes and 3 objects from each category.
CAMERA25 is a synthetic dataset that ontains the same categories as REAL275. It provides 300k synthetic RGB-D images of objects rendered on virtual background, with 25k images are withhold for testing.
HouseCat6D is a comprehensive multi-modal real-world dataset and encompasses 10 household categories.
The training set consists of 20k frames from 34 scenes and testing set consists of 3k frames across 5 scenes. With a total of 194 objects, each category contains 19 objects on average.

\noindent
{\bf Evaluation Metrics.} Following~\cite{chen2024secondpose,lin2024instance}, we evaluate the model performance with two metrics. (\textbf{i}) \textbf{3D IoU}. As for NOCS dataset, we report mean average precision (mAP) of CATRE~\cite{liu2022catre} Intersection over Union (IoU) with the thresholds of 75\%. For the HouseCat6D dataset, we report the mAP of 3D IoU under thresholds of 25\% and 50\%. (\textbf{ii}) \textbf{n°m \emph{cm}}.  We also utilize the combination of rotation and translation metrics of 5°2 \emph{cm}, 5°5 \emph{cm}, 10°2 \emph{cm} and 10°5 \emph{cm}, which means the estimation is considered correct when the error is below a threshold. 

\noindent
{\bf Implementation Details.} 
For a fair comparison, we utilize the same segmentation masks as AG-Pose~\cite{lin2024instance} and DPDN~\cite{lin2022category} from MaskRCNN~\cite{he2017mask} and resize them to $224 \times 224$. 
For model parameters, the feature dimensions are set as $C_1 = C_2 =$ 128, $C_3 =$ 768 and $C =$ 256, respectively. The number of point $N$ in point cloud is 1024 and the number of keypoints $N_{kpt}$ is set as 96. In dynamic queue, we set the size of queue $N_q$ to 80 and randomly select $N_{s} = 12$ features. 
For the use of pre-trained 3D models of ULIP-2, we apply PointBert~\cite{yu2022point} for knowledge distillation and employ PointNet++~\cite{qi2017pointnet++} for queue construction.
For the hyper-parameters setting, the balancing parameter $\mu$ in residual network is set as 0.1 following~\cite{huang2024froster}, and $\alpha_1$, $\alpha_2$ in overall loss function are 1 and 0.01, respectively.
For model optimizing, we employ the same data augmentation approach as previous works~\cite{lin2024instance,lin2022category}, which leverage random rotation degree sampled from $U$(0, 20) and rotation $\Delta t \sim U$(-0.02, 0.02) and scaling $\Delta s \sim U$(-0.18, 1.2).
The network is training on a single NVIDIA L40 GPU for a total of 120k iterations by the Adam~\cite{kingma2014adam} optimizer, with a mini training batch is 24 and a learning rate range from 2e-5 to 5e-4 based on triangular2 cyclical schedule~\cite{smith2017cyclical}.

\begin{table*}[htbp]
    \small
    \centering
    \setlength\tabcolsep{8pt}
    \begin{tabular}{l|c|c|c|cccc}
    \toprule
     Methods & Venue/Source & Shape Prior & $IoU_{75}^*$$\uparrow$ & 5°2\emph{cm}$\uparrow$ & 5°5\emph{cm}$\uparrow$ & 10°2\emph{cm}$\uparrow$ & 10°5\emph{cm}$\uparrow$ \\
    \midrule
    DPDN\cite{lin2022category}    &   ECCV'22  & \ding{51}   &54.0       & 46.0          &50.7         & 70.4       & 78.4       \\
    MH6D\cite{liu2024mh6d}    &   TNNLS'24   & \ding{51}     &54.2       & 53.0          &61.1         & 72.0       & 82.0      \\
    GCE-Pose\cite{li2025gce}    &   CVPR'25   & \ding{51}   &-       & {\ul57.0}         &{\ul65.1}         & {\ul75.6}       & \textbf{86.3}      \\
    \midrule
    HS-Pose\cite{zheng2023hs}    &   CVPR'23   & \ding{55}    &39.1       & 45.3          &54.9         & 68.6         & 83.6       \\
    VI-Net\cite{lin2023vi}    &   ICCV'23   & \ding{55}      &48.3       & 50.0          &57.6         & 70.8        & 82.1     \\
    CLIPose\cite{lin2024clipose}  &   TCSVT'24 & \ding{55}   & - & 48.5         & 58.2 & 70.3 & 85.1 \\
    GenPose\cite{zhang2023genpose}    &   NeurIPS'23  & \ding{55}  &-       & 52.1          &60.9         & 72.4        & 84.0      \\
    SecondPose\cite{chen2024secondpose}    &   CVPR'24   &\ding{55}     &49.7       & 56.2          &63.6         & 74.7       & 86.0     \\
    AG-Pose\cite{lin2024instance}    &   CVPR'24  & \ding{55}     &{\ul61.3}       & {\ul57.0}         &64.6         & 75.1       & 84.7      \\
    \midrule
    \textbf{CleanPose (ours)}    &   ~   & \ding{55}     & \textbf{62.7} & \textbf{61.7}         & \textbf{67.6} & \textbf{78.3} & \textbf{86.3}  \\
    \bottomrule
    \end{tabular}
    \vspace{-0.2cm}
    \caption{\textbf{Comparisons with state-of-the-art methods on REAL275 dataset.} $\uparrow$: a higher value indicating better performance. ‘*’ denotes CATRE~\cite{liu2022catre} IoU metrics and ‘-’ means unavailable statistics. Overall best results are in \textbf{bold} and the second best results are {\ul underlined}.
    }
    \vspace{-0.4cm}
    \label{tab:compare_sota}
\end{table*}
%
\begin{table}[htbp]
    \small
    \centering
    \setlength\tabcolsep{4pt}
    \begin{tabular}{l|c|ccccc}
    \toprule
   Methods & $IoU_{75}^*$  & 5°2\emph{cm} & 5°5\emph{cm} & 10°2\emph{cm}  & 10°5\emph{cm} \\
    \midrule
    HS-Pose\cite{zheng2023hs}               & -             & 73.3      & 80.5          & 80.4    & 89.4     \\
    CLIPose\cite{lin2024clipose}                  &-             & 74.8     & 82.2         & 82.0    &  91.2     \\
    GeoReF\cite{zheng2024georef}                  &79.2             & 77.9     & {\ul84.0}         & 83.8    & 90.5     \\
    AG-Pose\cite{lin2024instance}                  &\textbf{81.2}              & {\ul79.5}     & 83.7         & {\ul87.1}    & {\ul92.6}     \\
    \midrule
    \textbf{CleanPose (ours)}                   &{\ul80.7}             & \textbf{80.3}     & \textbf{84.2}         & \textbf{87.7}    &  \textbf{92.7}     \\
    \bottomrule
    \end{tabular}
    \vspace{-0.2cm}
    \caption{\textbf{Comparisons with state-of-the-art methods on CAMERA25 dataset.} A higher value indicating better performance. ‘*’ denotes CATRE~\cite{liu2022catre} IoU metrics and ‘-’ means unavailable statistics.  Overall best results are in \textbf{bold} and the second best results are {\ul underlined}.
    }
    \vspace{-0.2cm}
    \label{tab:compare_sota_camera}
\end{table}
\begin{table}[htbp]
    \small
    \centering
    \setlength\tabcolsep{1.8pt}
    \begin{tabular}{l|cc|cccc}
    \toprule
     Methods & $IoU_{25}$ & $IoU_{50}$ & 5°2\emph{cm} & 5°5\emph{cm} & 10°2\emph{cm} & 10°5\emph{cm} \\
    \midrule
    FS-Net\cite{chen2021fs}     &  74.9       & 48.0        & 3.3         & 4.2        & 17.1       & 21.6                     \\
    GPV-Pose\cite{di2022gpv}    &   74.9   & 50.7        & 3.5 & 4.6          & 17.8 & 22.7                                          \\
    VI-Net\cite{lin2023vi}    &   80.7        &56.4       & 8.4          &10.3         & 20.5        & 29.1     \\
    SecondPose\cite{chen2024secondpose}    &   83.7       &66.1       & 11.0          &13.4         & 25.3       & 35.7      \\
    AG-Pose\cite{lin2024instance}    &   {\ul88.1}      &{\ul76.9}      & {\ul21.3}         &{\ul22.1}         & {\ul51.3}       &{\ul54.3}      \\
    \midrule
    \textbf{CleanPose (ours)}    &  \textbf{89.2}        & \textbf{79.8}    & \textbf{22.4}         & \textbf{24.1} & \textbf{51.6} & \textbf{56.5}  \\
    \bottomrule
    \end{tabular}
    \vspace{-0.2cm}
    \caption{\textbf{Comparisons with prior-free state-of-the-art methods on HouseCat6D dataset.} A higher value indicating better performance. Overall best results are in \textbf{bold} and the second best results are {\ul underlined}.
    }
    \vspace{-0.4cm}
    \label{tab:compare_sota_housecat6d}
\end{table}

\subsection{Comparison with State-of-the-Art Methods}
\label{sec:compare_sota}
\begin{figure}[htbp]
\centering
\includegraphics[width=.9\columnwidth]{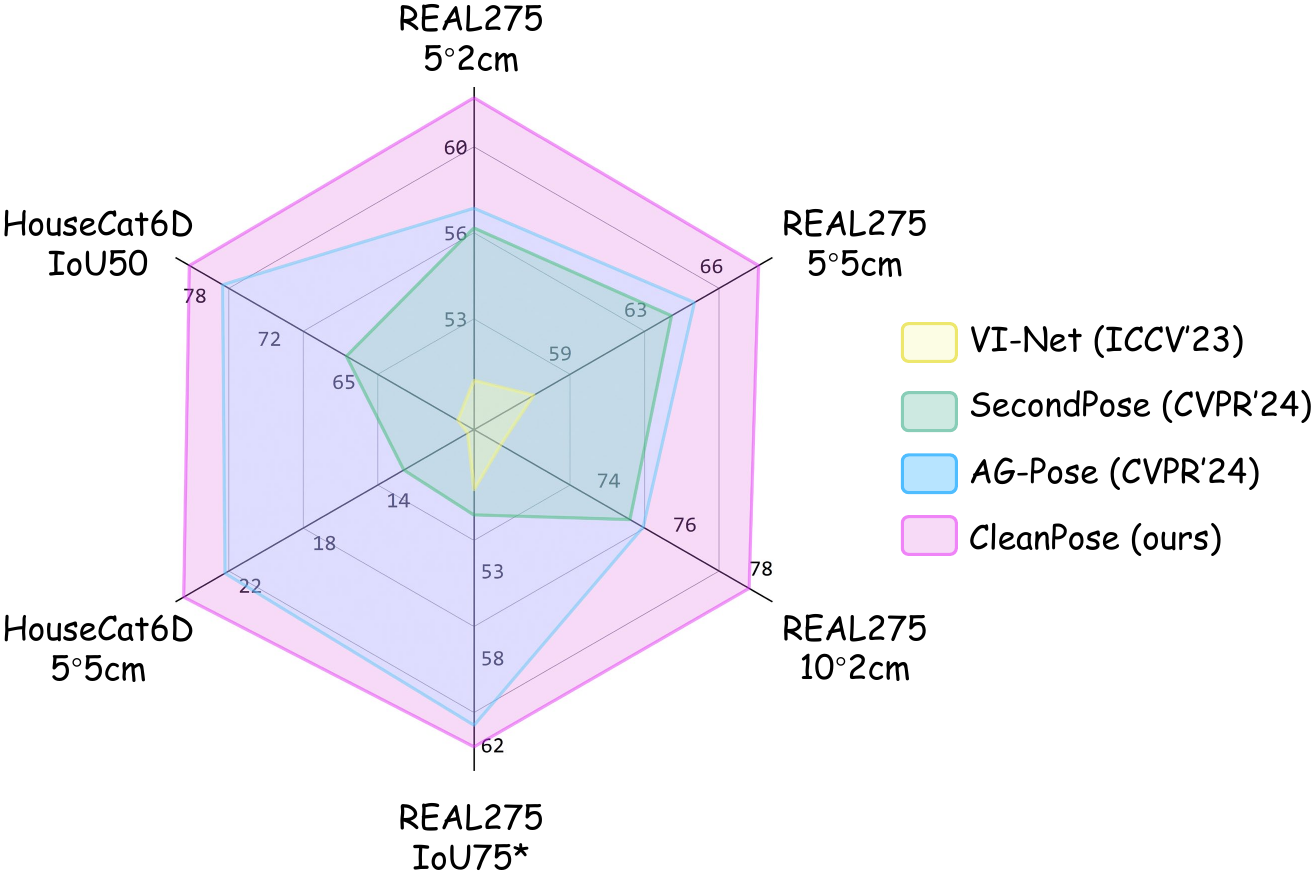}
   \caption{Comparison of ours with prior-free SOTA methods.
   }
   \vspace{-0.3cm}
   \label{fig:radar}
\end{figure}

As shown in \cref{fig:radar}, our method outperforms prior-free SOTA methods across key metrics in different datasets.

\begin{figure}[htbp]
\includegraphics[width=\columnwidth]{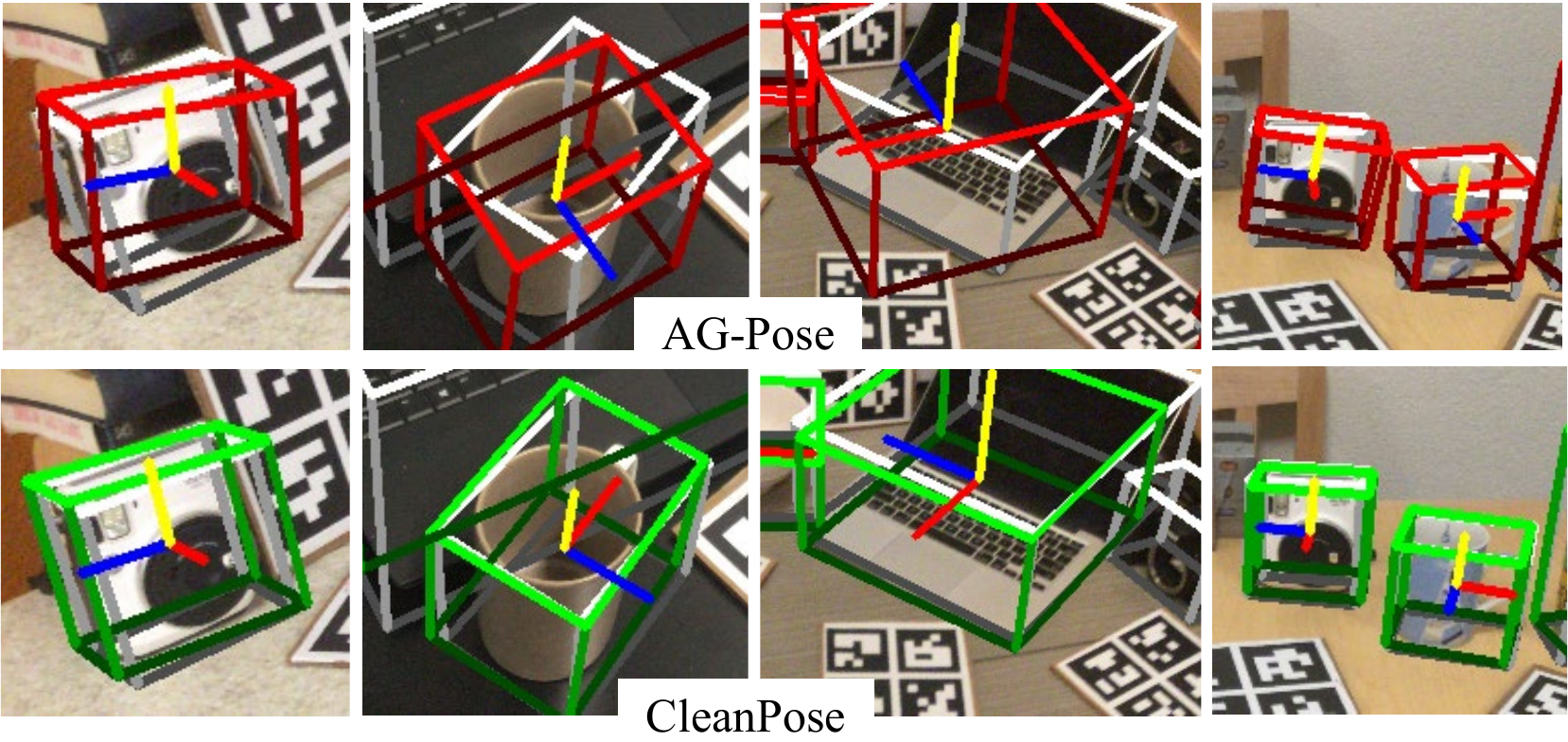}
   \caption{Qualitative comparison on REAL275~\cite{wang2019normalized}. We compare the predictions of CleanPose and the baseline AG-Pose~\cite{lin2024instance}. The ground truth is marked by white borders. 
   }
   \vspace{-0.4cm}
   \label{fig:vis}
\end{figure}

\noindent
{\bf Performance on REAL275 dataset.}
The comparisons between proposed CleanPose and previous methods on challenging REAL275 dataset are shown in \cref{tab:compare_sota}.
As can be easily observed, our CleanPose achieves the state-of-the-art performance in all metrics and outperforms all previous methods on REAL275 dataset.
Significantly, we achieve the precision of \textbf{61.7\%}, \textbf{67.6\%} and \textbf{78.3\%} in the rigorous metric of 5°2 \emph{cm}, 5°5 \emph{cm} and 10°2 \emph{cm}, surpassing the current prior-free state-of-the-art method AG-Pose~\cite{lin2024instance} with a large margin by 4.7\%, 3.0\% and 3.2\%, respectively.
Our method even outperforms prior-based sota method GCE-Pose~\cite{li2025gce} by 4.7\%, 2.5\% and 2.7\% in above three rigorous metric.
Moreover, the qualitative results of AG-Pose and proposed CleanPose are shown in \cref{fig:vis}. It can be observed that our method achieves significantly higher precision.
These exceptional outcomes further support the efficacy of our approach. Please refer to appendix for detailed comparison with SOTA method.


\noindent
{\bf Performance on CAMERA25 dataset.}
The comparison results are presented in \cref{tab:compare_sota_camera}. 
From the observation of results, it can be deduced that CleanPose ranks top performance across all the metrics except $IoU_{75}$, in which our method also achieves comparable performance with the best (80.7\% \vs 81.2\%). 
In detail, CleanPose achieves 80.3\% on 5°2 \emph{cm}, 84.2\% on 5°5 \emph{cm} and 87.7\% on 10°2 \emph{cm}, respectively. 
This superior performance on synthetic dataset further proves the effectiveness of CleanPose. 

\begin{table}[htbp]
    \small
    \centering
    \setlength\tabcolsep{3pt}
    \begin{tabular}{c|c|c|cccc}
    \toprule
   ID & Causal & Distillation & 5°2\emph{cm} & 5°5\emph{cm} & 10°2\emph{cm}  & 10°5\emph{cm} \\
    \midrule
    1        &\ding{55}      &  \ding{55}     &57.0          & 64.6    &75.1  &84.7   \\
    2        &\ding{51}      &  \ding{55}      & 59.7          &66.5    &77.5 &86.0   \\
    3         &\ding{55}     &\ding{51}        & 57.9        &65.5    &76.1 &85.2   \\
    \rowcolor{mygray}
    4         & \ding{51}             & \ding{51}      & \textbf{61.7}          & \textbf{67.6}   &\textbf{78.3} &\textbf{86.3}   \\
    \bottomrule
    \end{tabular}
    \vspace{-0.2cm}
    \caption{Effect of causal learning and knowledge distillation.
    }
    \vspace{-0.2cm}
    \label{tab:ablation_main}
\end{table}

\noindent
{\bf Performance on HouseCat6D dataset.}
In \cref{tab:compare_sota_housecat6d}, we evaluate our method on HouseCat6D~\cite{jung2024housecat6d} dataset.
The proposed CleanPose again achieves state-of-the-art performance in all metrics. In detail, our method outperforms the current best method AG-Pose~\cite{lin2024instance} by 2.9\% on $IoU_{50}$, 1.1\% on 5°2 \emph{cm}, 2.0\% on 5°5 \emph{cm} and 2.1\% on 10°5 \emph{cm}, respectively.
The overall and category-wise evaluation of 3D IoU on HouseCat6D dataset is provided in appendix.

\subsection{Ablation Studies}
\label{sec:ablation_studies}
We conduct ablation experiments to demonstrate the effectiveness of the proposed method on REAL275 dataset~\cite{wang2019normalized}.

\noindent
{\bf Effect of Causal Learning and Knowledge Distillation.} In \cref{tab:ablation_main}, we perform ablations of the proposed two main components. We adopt AG-Pose~\cite{lin2024instance} as our baseline, which servers as the current sota framework in COPE.
The results indicate that the integration of causal inference or knowledge distillation leads to significant enhancements in the model’s performance. This strongly demonstrates causal learning's considerable potential and the effectiveness of comprehensive category information guidance in COPE.

\noindent
{\bf Effect of different feature storage methods and update strategies.} In \cref{tab:ab_causal_data_update}, we investigate the impact of distinct combinations of feature storage and update strategies. ‘Similarity’ represents updating the closest features via similarity computation. 
The results indicate that employing the dynamic queue and FIFO update mechanism yields the best performance. 
We owe the advantage to the dynamic queue's superior ability to capture feature variations compared to a memory bank. Moreover, the FIFO ensures the removal of outdated features, which is beneficial for causal inference.


\noindent
{\bf Effect of adaptive fusion method.} As shown in \cref{tab:ab_causal_fusion}, our model achieves sota performance even without feature fusion. Introduction of the adaptive feature fusion method further improves the results in rigorous metrics.

\noindent
{\bf Effect of distinct category knowledge learning policies.} In \cref{tab:ab_dis_layer}, we assess the impact of category knowledge learning policies. ‘Concat’ means directly concatenating $\mathcal{F}^{ULIP}_{P}$ and $\mathcal{F}_{P}$, while ‘Contrastive’ indicates aligning above two features via contrastive learning. Our proposed residual distillation policy achieves the best performance since the residual layers effectively balance feature learning and category knowledge transfer.

\begin{table}[htbp]
    \small
    \centering
    \begin{subtable}[t]{1.0\linewidth}  
        \centering
        \setlength\tabcolsep{3pt}
        \begin{tabular}{cc|cccc}
            \toprule
            Data & Update & 5°2\emph{cm} & 5°5\emph{cm} & 10°2\emph{cm}  & 10°5\emph{cm}\\
            \midrule
            \rowcolor{mygray}
            Queue  & FIFO  & \textbf{61.7}          &\textbf{67.6}    &\textbf{78.3} &\textbf{86.3}    \\
            Queue & w/o update  & 57.1          &66.1    &77.5 &85.7   \\
            Queue & Similarity  & 61.0         &67.1   &77.6 &86.0   \\
            Memory bank & w/o update  & 57.6          &66.4    &75.6 &85.5   \\
            Memory bank & Similarity  &  59.1         &66.9    &77.5 &86.1   \\
            \bottomrule
        \end{tabular}
        \caption{Effect of different feature storage methods and update strategies.}
        \label{tab:ab_causal_data_update}
    \end{subtable}
    
     

     \vspace{0.2cm}  

    \begin{subtable}[t]{1.0\linewidth}  
        \centering
        \setlength\tabcolsep{10pt}
        \begin{tabular}{c|cccc}
            \toprule
            AF & 5°2\emph{cm} & 5°5\emph{cm} & 10°2\emph{cm}  & 10°5\emph{cm}\\
            \midrule
            \ding{55}  & 60.5          &66.5    &\textbf{78.8} &\textbf{86.5}   \\
            \rowcolor{mygray}
            \ding{51} & \textbf{61.7}          &\textbf{67.6}    &78.3 &86.3    \\
            \bottomrule
        \end{tabular}
        \caption{Effect of adaptive fusion in front-door adjustment.}
        \label{tab:ab_causal_fusion}
    \end{subtable}

    \vspace{0.2cm}  
    
    \begin{subtable}[t]{1.0\linewidth}  
        \centering
        \setlength\tabcolsep{7.5pt}
        \begin{tabular}{c|cccc}
            \toprule
            Policies & 5°2\emph{cm} & 5°5\emph{cm} & 10°2\emph{cm}  & 10°5\emph{cm}\\
            \midrule
            Concat & 60.0          &66.7    &77.9 &85.9   \\
            Contrastive & 59.8          &66.5    &77.6 &85.8   \\
            \rowcolor{mygray}
            Distillation & \textbf{61.7}          &\textbf{67.6}    &\textbf{78.3} &\textbf{86.3}   \\
            \bottomrule
        \end{tabular}
        \caption{Effect of distinct category knowledge learning policies.}
        \label{tab:ab_dis_layer}
    \end{subtable}

    \vspace{0.2cm}  

    \begin{subtable}[t]{1.0\linewidth}  
        \centering
        \setlength\tabcolsep{6pt}
        \begin{tabular}{c|cccc}
            \toprule
            3D Encoder & 5°2\emph{cm} & 5°5\emph{cm} & 10°2\emph{cm}  & 10°5\emph{cm}\\
            \midrule
            PointNet++\cite{qi2017pointnet++}  & 59.6          &66.8    &77.7 &85.8   \\
            PointMLP\cite{ma2022rethinking} & 60.8          &67.2    &77.4 &86.0   \\
            \rowcolor{mygray}
            PointBert\cite{yu2022point} & \textbf{61.7}          &\textbf{67.6}    &\textbf{78.3} &\textbf{86.3}   \\
            \bottomrule
        \end{tabular}
        \caption{Effect of different 3D encoders of ULIP-2~\cite{xue2024ulip} for distillation.}
        \label{tab:ab_dis_encoder}
    \end{subtable}
    \caption{Ablation studies on key details. Default settings are colored in \colorbox{mygray}{gray}.}
    \vspace{-0.3cm}
    \label{table:ablation_detail}
    
\end{table}

\noindent
{\bf Effect of different 3D encoders.} \cref{tab:ab_dis_encoder} illustrates the impact of using different 3D encoders in the distillation. Objectively, the pre-trained PointBert~\cite{yu2022point} realizes the sota point cloud zero-shot classification results~\cite{xue2024ulip}. Our model is also more performant using PointBert. Moreover, since our method selects PointNet++ as 3D encoder, we hypothesize that when the teacher and student models share similar architectures (\eg, both using PointNet++), the distillation may mislead the student model to focus on feature structure similarity rather than transferring category knowledge. 
Please refer to appendix for more ablations.
\section{Conclusion}
\label{sec:conclusion}
In this paper, we present CleanPose, the first solution that addresses the dataset biases in category-level pose estimation from the perspective of causal learning.
Motivated by pivotal observation that humans can learn inherent causality beyond biases, we formulate the modeling of crucial causal variables and develop a causal inference framework in COPE task. 
Furthermore, we devise a residual knowledge distillation network to transfer unbiased semantics knowledge from 3D foundation model, providing comprehensive causal guidance to achieve unbiased estimation.
Extensive experiments on challenging benchmarks demonstrate that CleanPose can significantly improve performance, showing the effectiveness of our method.
\section*{Acknowledgments}
This paper is supported by the National Natural Science Foundation of China under Grants (62233013, 62173248, 62333017, 624B2105).
\clearpage
\setcounter{page}{1}
\setcounter{section}{0}
\setcounter{table}{0}
\renewcommand\thetable{S\arabic{table}}
\setcounter{figure}{0}
\renewcommand\thefigure{S\arabic{figure}}
\maketitlesupplementary
\renewcommand\thesection{\Alph{section}}


\section{More Loss Function Details}
\label{sec:suppl_loss_details}
The backbone of our method is based on AG-Pose~\cite{lin2024instance}. In addition to $\mathcal{L}_{pose}$ (Eq. (\re{iccvblue}{21})), there are some additional loss functions to balance keypoints selection and pose prediction. First, to encourage the keypoints to focus on different parts, the diversity loss $\mathcal{L}_{div}$ is used to force the detected keypoints to be away from each other, in detail:
\begin{small} 
\begin{align}
    \mathcal{L}_{div}&=\sum_{i=1}^{N_{kpt}} \sum_{j=1, j \neq i}^{N_{kpt}} \mathbf{d}\left(\mathcal{P}_{kpt}^{(i)}, \mathcal{P}_{kpt}^{(j)}\right) \label{equ:supp_divloss1}\\
    \mathbf{d}\left(\mathcal{P}_{kpt}^{(i)}, \mathcal{P}_{kpt}^{(j)}\right)&=\max \left\{th_{1}-\left\|\mathcal{P}_{kpt}^{(i)}-\mathcal{P}_{kpt}^{(j)}\right\|_{2}, 0\right\} \label{equ:supp_divloss2},
\end{align}
\end{small}
where $th_1$ is a hyper-parameter and is set as 0.01, $\mathcal{P}_{kpt}^{(i)}$ means the $i$-th keypoint.
To encourage the keypoints to locate on the surface of the object and exclude outliers simultaneously, an object-aware chamfer distance loss $\mathcal{L}_{ocd}$ is employed to constrain the distribution of $\mathcal{P}_{kpt}$. In formula:
\begin{small} 
\begin{align}
    \mathcal{L}_{ocd}=\frac{1}{\left|\mathcal{P}_{kpt}\right|} \sum_{x_{i} \in \mathcal{P}_{kpt}} \min _{y_{j} \in \mathcal{P}_{obj}^{'}}\left\|x_{i}-y_{j}\right\|_{2}, \label{equ:supp_ocdloss}
\end{align}
\end{small}
where $\mathcal{P}_{obj}^{'}$ denotes the point cloud of objects without outlier points.
Moreover, we also use MLP to predict the NOCS coordinates of keypoints $\mathcal{P}_{kpt}^{nocs} \in \mathbb{R}^{N_{kpt} \times 3}$. Then, we generate ground truth NOCS of keypoints $\mathcal{P}_{kpt}^{gt}$ by projecting their coordinates under camera space $\mathcal{P}_{kpt}$ into NOCS using the ground truth $\mathcal{R}_{gt},t_{gt},s_{gt}$. And we use the $SmoothL_1$ loss to supervise the NOCS projection:
\begin{small} 
\begin{align}
    \mathcal{P}_{kpt}^{gt}&=\frac{1}{\left\|s_{gt}\right\|_{2}} \mathcal{R}_{gt}\left(\mathcal{P}_{kpt}-t_{gt}\right) \label{equ:supp_nocsgt}\\
    \mathcal{L}_{nocs} &= SmoothL_{1}(\mathcal{P}_{kpt}^{gt},\mathcal{P}_{kpt}^{nocs}). \label{equ:supp_nocsloss}
\end{align}
\end{small}
Hence, the complete form of overall loss (Eq. (\re{iccvblue}{22})) is as follows:
\begin{small} 
\begin{equation}
\begin{split}
    \mathcal{L}_{all} = \lambda_1\mathcal{L}_{ocd} + \lambda_2\mathcal{L}_{div} + \lambda_3\mathcal{L}_{nocs} \\ + \lambda_4\mathcal{L}_{pose} + \alpha_{2}\mathcal{L}_{KD}, \label{equ:supp_overall_loss}
\end{split}
\end{equation}
\end{small}
where the parameters are set as ($\lambda_1,\lambda_2,\lambda_3,\lambda_4,\alpha_{2}$) = ($1.0,5.0,1.0,0.3,0.01$) according to AG-Pose~\cite{lin2024instance} and following ablations.

\section{More Details of Using ULIP-2}
\label{sec:suppl_use_ulip2}
ULIP-2~\cite{xue2024ulip} is a large-scale 3D foundation model with strong perception capabilities for the point cloud modality. It offers multiple pre-trained versions of point cloud encoders. In our model, there are two key steps that involve the use of different pre-trained encoders of ULIP-2.
(\textbf{i}) In the knowledge distillation process, we leverage the pre-trained PointBert~\cite{yu2022point}, which achieves the best zero-shot classification performance across all versions. Therefore, it can provide comprehensive category knowledge guidance for our model. In the ablation study Tab. \re{iccvblue}{5d}, we also compared it with PointNet++~\cite{qi2017pointnet++}, which is more similar in architecture to our model. The experimental results demonstrate that our distillation method focuses more on category knowledge rather than feature similarity.
(\textbf{ii}) However, during the initial construction of the dynamic queue, we use the pre-trained PointNet++~\cite{qi2017pointnet++}, as the front-door adjustment primarily focuses on the differences between samples. We aim to avoid introducing confounders due to feature discrepancies from different encoders. The additional ablation study results in \cref{tab:supp_queue_encoder} also support our analysis.

\section{Additional Ablations}
\label{sec:suppl_additional_ablations}
\noindent
{\bf Effect of varying queue lengths $N_q$.} \cref{tab:supp_ab_causal_Nq} ablates the different lengths of dynamic queue $N_q$. The queue that is too short results in insufficient sample diversity, while too long affect memory efficiency and feature consistency.
We observe that the estimation performance achieves the peak at the length of around 80, with slight declines upon further increases.
We speculate that the queue length is closely related to task characteristics and data scale of COPE.
We select $N_q = 80$ in our model to balance between efficiency and accuracy.

\begin{table}[htbp]
    \small
    \centering
    \begin{subtable}[t]{1.0\linewidth}  
        \centering
        \setlength\tabcolsep{8pt}
        \begin{tabular}{c|cccc}
            \toprule
            $N_{q}$ & 5°2\emph{cm} & 5°5\emph{cm} & 10°2\emph{cm}  & 10°5\emph{cm}\\
            \midrule
            20  & 57.0          & 64.6    &75.1  &84.7   \\
            50 & 60.8          &67.3    &77.9 &\textbf{86.4}   \\
            \rowcolor{mygray}
            80 & \textbf{61.7}          &\textbf{67.6}    &\textbf{78.3} &86.3   \\
            200  &59.4          &66.1    &78.0 &85.9   \\
            500  &  58.8         &65.3    &76.8 &85.8   \\
            1000  & 58.3          &66.8    &76.3 &86.2   \\
            3000  & 57.7          &65.5    &75.6 &85.0   \\
            10000  & 57.0          &65.0    &75.7 &85.4   \\
            \bottomrule
        \end{tabular}
        \caption{Effect of varying queue lengths $N_{q}$}
        \label{tab:supp_ab_causal_Nq}
    \end{subtable}
    
    \vspace{0.2cm}  

    \begin{subtable}[t]{1.0\linewidth}  
        \centering
        \setlength\tabcolsep{9pt}
        \begin{tabular}{c|cccc}
            \toprule
            $N_{s}$ & 5°2\emph{cm} & 5°5\emph{cm} & 10°2\emph{cm}  & 10°5\emph{cm}\\
            \midrule
            6  & 60.7          & 66.3    &77.8  &85.8   \\
            \rowcolor{mygray}
            12 & \textbf{61.7}          &\textbf{67.6}    &78.3 &86.3   \\
            18 & 59.4          &66.5    &\textbf{78.8} &\textbf{86.8}   \\
            24  &  58.5         &65.8    &77.8 &85.9   \\
            48  &56.8          &64.9    &76.3 &85.6   \\
            80  & 56.9          &64.4    &76.2 &85.8   \\
            \bottomrule
        \end{tabular}
        \caption{Effect of varying queue lengths $N_{s}$}
        \label{tab:supp_ab_causal_Ns}
    \end{subtable}

    \vspace{0.2cm}  

    \begin{subtable}[t]{1.0\linewidth}  
        \centering
        \setlength\tabcolsep{8pt}
        \begin{tabular}{c|cccc}
            \toprule
            $\alpha_{2}$ & 5°2\emph{cm} & 5°5\emph{cm} & 10°2\emph{cm}  & 10°5\emph{cm}\\
            \midrule
            0.005  & 59.3          & 66.8    &78.0  &\textbf{86.4}   \\
            \rowcolor{mygray}
            0.01 & \textbf{61.7}          &\textbf{67.6}    &\textbf{78.3} &86.3   \\
            0.1 & 58.4          &65.1    &77.6 &85.9   \\
            0.5  &  57.3         &63.9    &76.2 &86.0   \\
            1  &  56.9         &63.4    &76.4 &85.4   \\
            \bottomrule
        \end{tabular}
        \caption{Effect of varying balanced coefficient $\alpha_2$}
        \label{tab:supp_ab_kd_a2}
    \end{subtable}
     
    \caption{Additional ablation studies on some hyper-parameters. Default settings are colored in \colorbox{mygray}{gray}.}
    \label{table:supp_ablation_detail}
    \vspace{-0.3cm}
\end{table}

\vspace{0.1cm}
\noindent
{\bf Effect of different sampling quantities $N_s$.} In Sec. \re{iccvblue}{4.2} of main manuscript, we sample $N_s$ features for the specific network design to perform \emph{front-door adjustment}. We study the influence with response to different sampling quantities $N_s$ in \cref{tab:supp_ab_causal_Ns}. It can be found that a large $N_s$ leads to a slight performance degradation. We speculate that a larger sample size may introduce noise and redundant information that affects key features in causal inference. An appropriate sample size can balance valid and redundant information, prompting the model to focus on learning more representative causal correlations. The results demonstrate that $N_s = 12$ yields the most significant performance gains.

\vspace{0.1cm}
\noindent
{\bf Varying balanced coefficient $\alpha_{2}$ for loss $\mathcal{L}_{KD}$.} In Sec. \re{iccvblue}{4.3} of main manuscript, we introduce L2 loss to supervise the feature-based distillation and use $\alpha_{2}$ to balanced its contribution in overall loss function. We investigate the impact of different $\alpha_{2}$ in \cref{tab:supp_ab_kd_a2}. We observe that the better performance is achieved when $\alpha_{2}$ is small, possibly because $\mathcal{L}_{KD}$ is comparable in magnitude to the pose loss function, which is favorable for regression. The results show that our method performs well under $\alpha_{2} = 0.01$.

\vspace{0.1cm}
\noindent
{\bf Different queue initialization approaches.} By default, we construct confounders queue with features extracted by 3D encoders of ULIP-2~\cite{xue2024ulip}. Alternatively, we can randomly initialize the queue, which should achieve the same effect ideally. Therefore, we evaluate the performance between two initialization approaches in \cref{tab:supp_ab_queue_init}. 
The results indicate the degraded performance with “Random” initialization strategy. We speculate that the randomly initialized queue may introduce additional and uncontrollable confounders, limiting the model's optimization potential.

\vspace{0.1cm}
\noindent
{\bf Effect of different 3D encoder for initial construction of the queue.}
\cref{tab:supp_queue_encoder} ablates the different 3D encoders of ULIP-2~\cite{xue2024ulip} for initial construction of the dynamic queue. The results exhibit that using PointNet++~\cite{qi2017pointnet++} yields the most performance gains.
As mentioned in \cref{sec:suppl_use_ulip2}, the dynamic queue is utilized in the cross-attention phase of front-door adjustment, thus primarily focusing on the differences between samples. Using encoders with similar architectures helps avoid introducing extra confounders.

\begin{table}[htbp]
    \small
    \centering
    \begin{subtable}[t]{1.0\linewidth}  
        \centering
        \setlength\tabcolsep{9pt}
        \begin{tabular}{c|cccc}
            \toprule
            Init. & 5°2\emph{cm} & 5°5\emph{cm} & 10°2\emph{cm}  & 10°5\emph{cm}\\
            \midrule
            Random  & 58.0          & 65.7    &75.8  &85.1   \\
            \rowcolor{mygray}
            Extract & \textbf{61.7}   &\textbf{67.6}    &\textbf{78.3} &\textbf{86.3}   \\
            \bottomrule
        \end{tabular}
        \caption{Effect of different queue initialization approaches}
        \label{tab:supp_ab_queue_init}
    \end{subtable}

    \vspace{0.2cm}  

    \begin{subtable}[t]{1.0\linewidth}  
        \centering
        \setlength\tabcolsep{7pt}
        \begin{tabular}{c|cccc}
            \toprule
            3D Encoder & 5°2\emph{cm} & 5°5\emph{cm} & 10°2\emph{cm}  & 10°5\emph{cm}\\
            \midrule
            \rowcolor{mygray}
            PointNet++\cite{qi2017pointnet++} & \textbf{61.7}          &\textbf{67.6}    &\textbf{78.3} &86.3   \\
            PointMLP\cite{ma2022rethinking} & 58.1          &65.8   &76.3 &85.2   \\
            PointBert\cite{yu2022point} & 58.8          &65.6    &77.9 &\textbf{86.4}   \\
            \bottomrule
        \end{tabular}
        \caption{Effect of different 3D encoders of ULIP-2~\cite{xue2024ulip} for initial construction of the queue.}
        \label{tab:supp_queue_encoder}
    \end{subtable}

    \vspace{0.2cm}  

    \begin{subtable}[t]{1.0\linewidth}  
        \centering
        \setlength\tabcolsep{7pt}
        \begin{tabular}{c|cccc}
            \toprule
            Selector & 5°2\emph{cm} & 5°5\emph{cm} & 10°2\emph{cm}  & 10°5\emph{cm}\\
            \midrule
            \rowcolor{mygray}
            Random  & \textbf{61.7}          &\textbf{67.6}    &\textbf{78.3} &86.3   \\
            K-means & 58.5          &65.5    &78.0 &\textbf{86.5}   \\
            K-means (simi) & 58.7          &66.2    &77.7 &86.0   \\
            \bottomrule
        \end{tabular}
        \caption{Effect of distinct feature selectors}
        \label{tab:supp_queue_sample}
    \end{subtable}

    \caption{Additional ablation studies on confounders queue. Default settings are colored in \colorbox{mygray}{gray}.}
    \label{table:supp_ablation_detail}
\end{table}

\begin{figure}[htbp]
    \centering
    \subfloat[Random feature selector]
    {\includegraphics[width=.4\columnwidth]{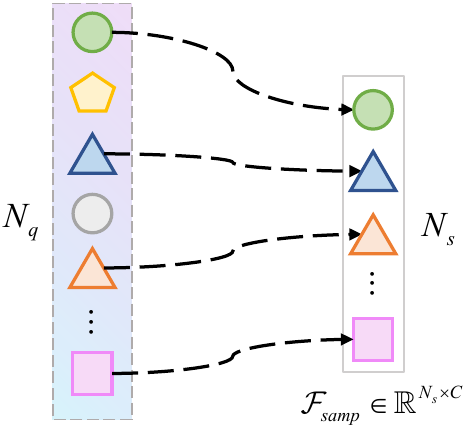}}
    \hspace{8pt}
    \subfloat[K-means feature selector]{\includegraphics[width=.51\columnwidth]{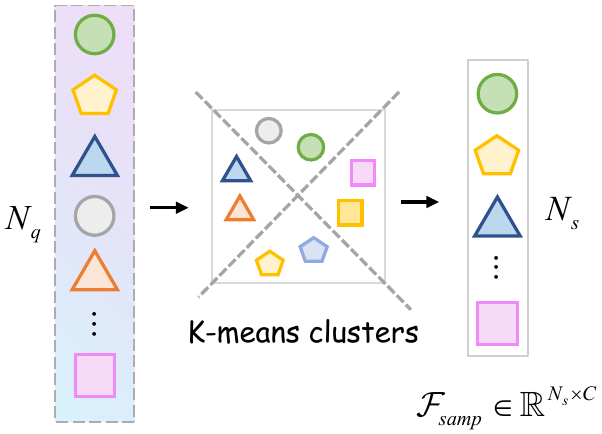}}
    \caption{Illustration of different feature selectors in ablations \cref{tab:supp_queue_sample}.
    }
    \label{fig:subfig}
    \vspace{-0.2cm}
\end{figure}

\vspace{0.1cm}
\noindent
{\bf Various feature selection strategies.} In Sec. \re{iccvblue}{4.2} of main manuscript, we randomly sample $N_s$ features from queue by default, as shown in \cref{fig:subfig}(a). Optionally, we can first use K-means to cluster the features of the queue, and then select features from each cluster to form $\mathcal{F}_{samp}$, as illustrated in \cref{fig:subfig}(b). For fair comparison, the number of clusters is set equal to $N_s$. 
We investigate the impact of these two feature selector in \cref{tab:supp_queue_sample}. As shown in the table, we observe that K-means-based feature sampling strategy shows a decline in performance on strict metrics (5°2\emph{cm} and 5°5\emph{cm}). We argue that k-means, which clusters by Euclidean distance, may lose important boundary information, thus affecting the model performance. Moreover, k-means clustering needs to be performed after each queue update, which increases computational load and training costs. Therefore, we added a comparative experiment using similarity-based updates, denoted as ‘K-means (simi)’, where (simi) refers to the ‘Similarity’ defined in Tab. \re{iccvblue}{5a} of main text. 
\begin{table*}[htbp]
    \small
    \centering
    \setlength\tabcolsep{7pt}
    \begin{tabular}{l|ccc|cccc}
    \toprule
     Category & $IoU_{25}^*$$\uparrow$ & $IoU_{50}^*$$\uparrow$   & $IoU_{75}^*$$\uparrow$ & 5°2\emph{cm}$\uparrow$ & 5°5\emph{cm}$\uparrow$ & 10°2\emph{cm}$\uparrow$  & 10°5\emph{cm}$\uparrow$ \\ 
     \midrule
    bottle &51.3 &49.4 &37.1    &75.7 &81.7 &79.9 &87.8 \\
    bowl &100.0 &100.0 &93.8   &93.3 &98.2 &95.0 &99.9  \\
    camera &90.9 &83.5 &39.9   &2.8 &3.2 &33.9 &40.5  \\
    can &71.3 &71.1 &43.2      &84.2 &85.9 &96.9 &98.6  \\
    laptop &86.3 &84.0 &76.1      &69.2 &90.9 &71.9 &98.5  \\
    mug &99.6 &99.4 &86.1      &45.2 &45.6 &91.9 &91.9  \\
    \midrule
    Average &83.3 &81.2 &62.7 &61.7 &67.6 &78.3 &86.3 \\
    \bottomrule
    \end{tabular}
    \caption{\textbf{Category-wise evaluation of CleanPose on REAL275 dataset.} ‘*’ denotes CATRE~\cite{liu2022catre} IoU metrics.
    }
    \label{tab:suppl_per_category_real}
\end{table*}
\begin{table*}[htbp]
    \footnotesize
    \centering
    \setlength\tabcolsep{4pt}
    \begin{tabular}{l|c|c|cccccccccc}
    \toprule
     \multirow{2}{*}{Methods} & \multirow{2}{*}{$IoU_{75}$$\uparrow$} & \multicolumn{11}{c}{$IoU_{25}$/$IoU_{50}$$\uparrow$}\\
     \cline{3-13}
     ~ & ~ & Average    & Bottle & Box & Can & Cup  & Remote & Teapot &Cutlery & Glass & Tube & Shoe\\ 
     \midrule
    NOCS\cite{wang2019normalized} &-& 50.0/21.2 & 41.9/5.0  & 43.3/6.5  & 81.9/62.4 & 68.8/2.0  & \textbf{81.8}/\textbf{59.8} & 24.3/0.1  & 14.7/6.0  & 95.4/49.6 & 21.0/4.6  & 26.4/16.5 \\
    FS-Net\cite{chen2021fs} &14.8 &74.9/48.0 & 65.3/45.0 & 31.7/1.2  & 98.3/73.8 & 96.4/68.1 & 65.6/46.8 & 69.9/59.8 & 71.0/51.6 & 99.4/32.4 & 79.7/{\ul46.0} & 71.4/55.4 \\
    GPV-Pose\cite{di2022gpv} &15.2 &74.9/50.7 & 66.8/45.6 & 31.4/1.1  & 98.6/75.2 & 96.7/69.0 & 65.7/46.9 & 75.4/61.6 & 70.9/52.0 & {\ul99.6}/62.7 & 76.9/42.4 & 67.4/50.2 \\
    VI-Net\cite{lin2023vi} &20.4 &80.7/56.4 & 90.6/79.6 & 44.8/12.7 & {\ul99.0}/67.0 & 96.7/72.1 & 54.9/17.1 & 52.6/47.3 & 89.2/76.4 & 99.1/93.7 & \textbf{94.9}/36.0 & 85.2/62.4 \\
    SecondPose\cite{chen2024secondpose} &24.9 &83.7/66.1 & 94.5/{\ul79.8} & \textbf{54.5}/{\ul23.7} & 98.5/93.2 & {\ul99.8}/{\ul82.9} & 53.6/35.4 & 81.0/71.0 & 93.5/74.4 & 99.3/92.5  & 75.6/35.6 & 86.9/73.0 \\
    AG-Pose\cite{lin2024instance} &{\ul53.0} &{\ul88.1}/{\ul76.9} & {\ul97.6}/\textbf{86.0}  & {\ul54.0}/13.9   & 98.3/{\ul96.7} & \textbf{100}/\textbf{99.9}  & 53.9/37.2 & \textbf{99.9}/\textbf{98.5} & {\ul96.0}/\textbf{93.3}   & \textbf{100}/{\ul99.3}  & {\ul81.4}/45.0   & {\ul99.7}/{\ul99.5} \\
    \midrule
    \textbf{CleanPose} &\textbf{53.9} &\textbf{89.2}/\textbf{79.8} & \textbf{99.9}/79.1 & 51.4/\textbf{28.7} & \textbf{99.9}/\textbf{99.7} & \textbf{100}/\textbf{99.9}  & {\ul71.2}/{\ul57.8} & {\ul99.0}/{\ul94.0} & \textbf{97.8}/{\ul91.0} & \textbf{100}/\textbf{99.6}  & 72.7/\textbf{48.4} & \textbf{99.8}/\textbf{99.8}\\
    \bottomrule
    \end{tabular}
    \caption{\textbf{Overall and category-wise evaluation of 3D IoU on the HouseCat6D.} $\uparrow$: a higher value indicating better performance, ‘-’ means unavailable statistics. Overall best results are in \textbf{bold} and the second best results are {\ul underlined}.
    }
    \label{tab:suppl_housecat6d}
\end{table*}
\begin{table}[htbp]
    \small
    \centering
    \setlength\tabcolsep{5pt}
    \begin{tabular}{c|cccccc|c}
    \toprule
   Seed & 1 & 42 & 500 & 1k & 1w  & 10w& $\sigma^{2}\downarrow$ \\
    \midrule
    5°2\emph{cm}              &  61.4     &61.5         & 61.7    &61.4  &61.3 &61.7&0.03   \\
    5°5\emph{cm}     &67.2        & 67.3        &67.5    &	67.2 &67.1 &67.6&0.04  \\
    \bottomrule
    \end{tabular}
    \caption{Effect of different sampling seed during inference.
    }
    \vspace{-0.2cm}
    \label{tab:inference_samp}
\end{table}
In this case, clustering is only performed once during the initial training. However, experimental results also show that such strategy leads to further performance degradation as one clustering loses the diversity of features.

\vspace{0.1cm}
\noindent
{\bf Effect of different sampling seed during inference.} We have conducted additional experiments with 6 different random seeds, as shown in \cref{tab:inference_samp}. The computed variances $\sigma^{2}$ for metrics demonstrate stable performance across different random seeds, indicating the robustness and reliability of our method.


\section{More Experimental Results}
\label{sec:suppl_more_results}
We report category-wise results of REAL275~\cite{wang2019normalized} in \cref{tab:suppl_per_category_real}. Since there is a small mistake in the original evaluation code of NOCS~\cite{wang2019normalized} for the 3D IoU metrics, we present more reasonable CATRE~\cite{liu2022catre} metrics following~\cite{zheng2024georef,liu2024mh6d,chen2024secondpose}. 
Further, more detailed results of HouseCat6D~\cite{jung2024housecat6d} are shown in \cref{tab:suppl_housecat6d}. As for more restricted metric $IoU_{75}$, our method also demonstrates state-of-the-art performance (\textbf{53.9\%}), further validating the effectiveness of CleanPose in 3D IoU evaluation.
\begin{table}[htbp]
    \small
    \centering
    \setlength\tabcolsep{2pt}
    \begin{tabular}{c|l|c|c|c|c|c|c}
    \toprule
   ID & Method & Visual Enc. & Param.$\downarrow$ & $IoU_{75}^* \uparrow$ & 5°2\emph{cm}$\uparrow$ & TT$\downarrow$  & IT$\uparrow$\\
    \midrule
    1        &AG-Pose      &  ViT-S/14     &\textbf{223M}     &61.3         &57.0 &\textbf{51.3} &\textbf{35}   \\
    \rowcolor{mygray}
    2         &ours     &ViT-S/14        & \underline{246M}     &\textbf{62.7}       &	\textbf{61.7} &51.8 &\underline{33}   \\
    3        &ours$^{\dag}$      &  ViT-S/14      & \underline{246M}     &61.8         &58.1 &\underline{51.6} &\underline{33}   \\
    \midrule
    4         & AG-Pose        & ResNet18      & \textbf{220M}      &60.9       &56.2 &\textbf{51.2} &\textbf{35}   \\
    5         & ours        & ResNet18      & \underline{243M}      &\underline{62.3}      &\underline{60.3} &\underline{51.6} &\underline{33}   \\
    \bottomrule
    \end{tabular}
    \caption{Detailed comparison results. ‘*’ denotes CATRE~\cite{liu2022catre} IoU metrics. ‘$\dag$’ represents replacement of causal module with MLPs of the same number of parameters. TT: Traning Time (min/epoch), IT: Inference Speed (Frame/sec). Overall best results are in \textbf{bold} and the second best results are {\ul underlined}. Default settings are colored in \colorbox{mygray}{gray}.
    }
    \vspace{-0.2cm}
    \label{tab:ablation_comparison}
\end{table}
Moreover, in category-wise validation on $IoU_{25}$ and $IoU_{50}$, our approach obtains state-of-the-art (\eg, \emph{Can}, \emph{Cup}, \emph{Glass} and \emph{Shoe}) or competitive results across all categories. 
It is worth mentioning that our method exhibits more stable performance on these two metrics. For instance, compared to the current sota method AG-Pose~\cite{lin2024instance} in the \emph{Box} category, our method achieves the best performance (\textbf{28.7\%}) on $IoU_{50}$ metric when both obtain competitive results on $IoU_{25}$ metric, with a significant reduction of the AG-Pose (13.9\%).

\begin{figure}[htbp]
\centering
\includegraphics[width=\columnwidth]{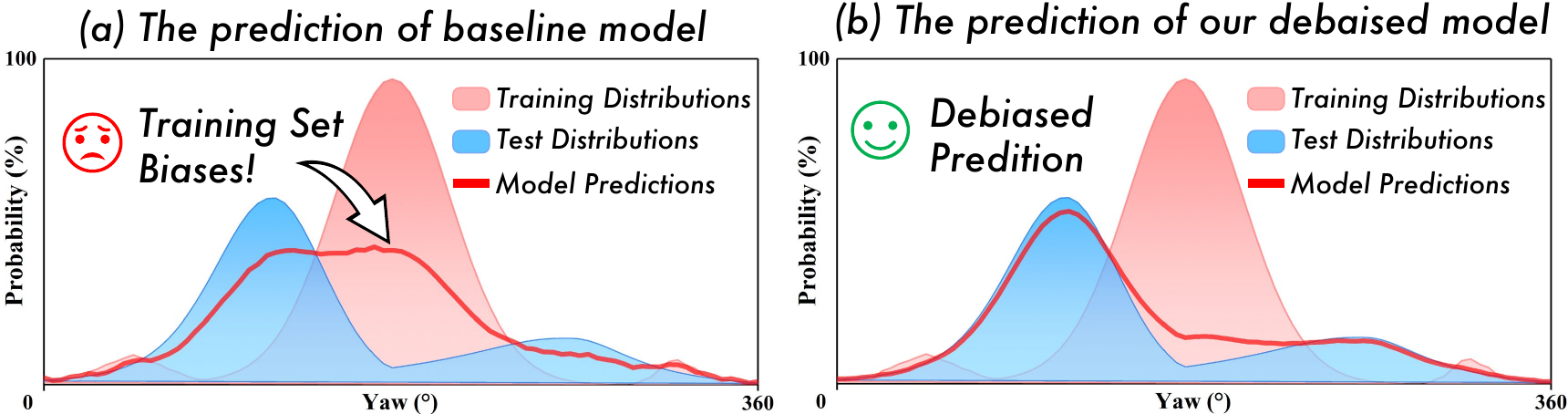}
   \caption{Effect of debiasing. Illustrated by Yaw angle distributions of the \emph{mug} category.
   }
   \vspace{-0.4cm}
   \label{fig:debiased}
\end{figure}

\section{Detailed Comparison with SOTA method}
\label{sec:suppl_detailed_comparison}
We follow the domain consensus to report the metric accuracy in main manuscript. Moreover, we add more terms, \eg visual encoder type, inference latency (FPS), in \cref{tab:ablation_comparison} for comprehensive comparison.
As confirmed in (\#1) and (\#2) of \cref{tab:ablation_comparison}, the front-door adjustment only increases the number of parameters by {\bf 10\%} (246M \vs 223M), while the running time remains nearly unchanged (33 \vs 35 in FPS). To ensure a fair comparison, we also replace the front-door module with MLPs that have the same number of parameters (\#3). The results further demonstrate the superior effectiveness of causal learning.
What's more, we include additional results with ResNet18. Specifically, our method still outperforms AG-Pose with ResNet18 setting (\#4 \vs \#5), further supporting the efficacy of our approach.

\vspace{0.1cm}
\section{Effect of Debiasing.}
\label{sec:suppl_Debiasing}
\noindent
We evaluate the debiasing effect via rotation distributions. As shown in \cref{fig:debiased}, the predictions of baseline model are clearly biased toward the training set distributions, while the debiased model primarily unaffected.

\section{Limitation and Broader Impact}
\label{sec:suppl_limitation}
\noindent
{\bf Limitation and future work.} While our method achieves superior results in various challenging benchmarks of category-level pose estimation, there are still several aspects for improvement. First, although the front-door adjustment is effective, the investigate on the application of causal learning methods remains incomplete. Therefore, exploring further use of different causal learning methods such as back-door adjustment and counterfactual reasoning may enhance the performance of CleanPose.
Second, despite the guidance of the causal analysis, the network modules in actual implementation may induce inaccuracy inevitably. Such a flaw introduces a gap between the causal framework and the network design. In future work, we will further study advanced algorithm design strategies.

\vspace{0.1cm}
\noindent
{\bf Broader Impact.} For tasks with parameter regression properties, \eg, category-level pose estimation, the current mainstream approaches focus on exploring advanced network designs to perform data fitting. We believe that relying solely on learning statistical similarity can also introduce spurious correlations into parameter regression models, thereby damaging the model's generalization ability. We hope this work brings new insights for the broader and long-term research on parameter regression tasks.
Besides, adapting foundation models to downstream tasks has become a dominant paradigm in machine learning. Our method also provide novel views for offering knowledge guidance in similar tasks across diverse categories.


{
    \small
    \bibliographystyle{ieeenat_fullname}
    \bibliography{main}
}


\end{document}